%% file: main.tex
\documentclass[letterpaper]{article} % DO NOT CHANGE THIS
\usepackage{aaai25}  % DO NOT CHANGE THIS
\usepackage{times}  % DO NOT CHANGE THIS
\usepackage{helvet}  % DO NOT CHANGE THIS
\usepackage{courier}  % DO NOT CHANGE THIS
\usepackage[hyphens]{url}  % DO NOT CHANGE THIS
\usepackage{graphicx} % DO NOT CHANGE THIS
\usepackage{natbib}  % DO NOT CHANGE THIS AND DO NOT ADD ANY OPTIONS TO IT
\usepackage{caption} % DO NOT CHANGE THIS AND DO NOT ADD ANY OPTIONS TO IT
\usepackage{algorithm}
\usepackage{algorithmic}

\usepackage{newfloat}
\usepackage{listings}
\DeclareCaptionStyle{ruled}{labelfont=normalfont,labelsep=colon,strut=off} % DO NOT CHANGE THIS
\floatstyle{ruled}
\newfloat{listing}{tb}{lst}{}
\floatname{listing}{Listing}

\usepackage{amsmath}
\usepackage{epigraph}
\usepackage{tikz}
\usetikzlibrary{arrows}
\usetikzlibrary {arrows.meta, bending}
\usepackage{csquotes}
\usepackage{xcolor}
\usepackage{amssymb}

\newtheorem{example}{Example}%[section]
\newtheorem{definition}{Definition}[section]
\newtheorem{remark}{Remark}[section]

\title{Toward A Causal Framework for Modeling Perception}
\author {
    Jose M. Alvarez\textsuperscript{\rm 1},
    Salvatore Ruggieri\textsuperscript{\rm 2}
}
\affiliations {
    \textsuperscript{\rm 1}Department of Computer Science, KU Leuven\\
    \textsuperscript{\rm 2}Department of Computer Science, University of Pisa\\
    josemanuel.alvarez@kuleuven.be, salvatore.ruggieri@unipi.it
}

\begin{document}

\maketitle

\begin{abstract}
Perception occurs when individuals interpret the same information differently. It is a known cognitive phenomenon with implications for bias in human decision-making. Perception, however, remains understudied in machine learning (ML). This is problematic as modern decision flows, whether partially or fully automated by ML applications, always involve human experts. For instance, how might we account for cases in which two experts interpret differently the same deferred instance or explanation from a ML model? Addressing this and similar questions requires first a formulation of perception, particularly, in a manner that integrates with ML-enabled decision flows. In this work, we present a first approach to modeling perception causally. We define perception under causal reasoning using structural causal models (SCMs). Our approach formalizes individual experience as additional causal knowledge that comes with and is used by the expert decision-maker in the form of a SCM. We define two kinds of probabilistic causal perception: structural and parametrical. We showcase our framework through a series of examples of modern decision flows. We also emphasize the importance of addressing perception in fair ML, discussing relevant fairness implications and possible applications.
\end{abstract}

%     % \link{Code}{https://aaai.org/example/code}
%     % \link{Datasets}{https://aaai.org/example/datasets}
%     \link{Extended version}{https://arxiv.org/abs/2401.13408}

\section{Introduction}
\label{sec:Introduction}
\input{sections/1_Introduction}

\section{Background}
\label{sec:Introduction.Background}
\input{sections/2_Background}

\section{Perception of Probability}
\label{sec:ProbabilisticPerception}
\input{sections/3_ProbabilisticPerception}

\section{Formulating Causal Perception}
\label{sec:CausalPerception}
\input{sections/4_CausalPerception}

\section{Toward a Causal Modeling Framework}
\label{sec:TheFramework}
\input{sections/5_TheFramework}

\section{Perception and Fairness}
\label{sec:Fairness}
\input{sections/6_PerceptionAndFairness}

\section{Conclusion}
\label{sec:Discussion}
\input{sections/7_Discussion}

\appendix

\section{Ethical Statement}
We did not face ethical challenges when drafting this paper.

\section{Acknowledgments}
We are indebted to Riccardo Massida for his discussions, insights, and suggestions. 
We would like to thank also Mayra Russo, Maria-Esther Vidal, and Andrea Pugnana for their feedback in earlier versions of this paper.
Salvatore Ruggieri is supported by the European Union (EU)’s Horizon Europe research and innovation program for the project FINDHR (g.a. 101070212).
% , and under the Horizon 2020 Marie Sklodowska-Curie Actions research and innovation program  for the project NoBIAS (g.a. 860630). 
Views and opinions expressed are those of the authors only and do not necessarily reflect those of the EU. Neither the EU nor the granting authority can be held responsible for them.

\bibliography{references}

\end{document}

%% file: sections/1_Introduction.tex
The same piece of information can be interpreted differently by individuals.  
The late \citet{KahnemanNIPS2021} refers to this cognitive phenomenon as perception.
It is a product of the mental heuristics humans use to enable fast decision-making under uncertainty.
These heuristics and, in turn, perception are shaped by individual experience, which can lead to biased decision-making.
The Linda Problem \cite{tversky_judgment_1974, tversky_judgments_1981, tversky_extensional_1983}, for instance, illustrates how perception overrides logical reasoning when individuals rely on intuitive judgment. 
In a series of famous experiments, participants committed the conjunction fallacy by judging the conjunction (``Linda is a bank teller and a feminist'') as more probable than a single constituent (``Linda is a bank teller"), violating the basic laws of probability that govern rational decision-making.
These and similar decision-making experiments \cite{Kahneman2011Thinking, Kahneman2016Noise, Thaler2008Nudge}, together with the growing regulatory emphasis on trustworthy machine learning (ML) applications  \cite{DBLP:journals/ethicsit/AlvarezCEFFFGMPLRSSZR24}, underscore the need to consider perception as a core problem in ML.

Modern decision processes are powered by ML applications and always involve some degree of human decision-making \cite{DBLP:conf/aaai/Ruggieri0PST23, Scantamburlo2024_PredictionAndDecision, DBLP:conf/aaai/RuggieriP25}.
In learning to defer (LtD), for example, the goal is to learn a ML model that abstains from predicting on instances it is not certain of and defers the decision to a human expert \cite{DBLP:conf/nips/MadrasPZ18}.
In explainable artificial intelligence (xAI), for example, the goal is to develop methods that explain to an expert the predictions of a ML model \cite{DBLP:journals/expert/GuidottiMGPRT19}.
Perception can clearly occur in both settings when multiple individual experts are involved, potentially impacting the outputs of these ML applications.

How might we develop reliable LtD and xAI methods when experts may disagree on the same deferred instance or explanation?
These questions are often ignored when building modern decision flows (see, e.g., \citet{DBLP:conf/aaai/CabitzaCB23, DBLP:journals/pacmhci/FahimiRSVBK24, DBLP:conf/kdd/SrivastavaHK19}).
Importantly, addressing this gap requires a formalization of perception that aligns with the standard ML problem formulation, which is probabilistic \cite{Pearl1988ProbabilisticReasoning, PearlCausality2009, Goodfellow2016DeepLearning}.
In this work, we provide a first approach to this objective.

\subsection{Our Contributions}
\label{sec:Introduction.Contributions}

How can we formally represent individuals who disagree in their interpretation of the same information?
We first propose a probabilistic framework for perception in which individuals construct disagreeing probability distributions, $P(\mathbf{X})$, for the same information $\mathbf{X}$.
Under this framework, for example, we can explain why one expert assigns a higher probability to a positive classification for a deferred instance compared to another expert.
We then propose causal perception by extending the initial formulation to account for probabilistic causal reasoning.
Here, individuals are equipped with structural causal models (SCMs) \cite{PearlCausality2009} that not only describe $P(\mathbf{X})$ but also govern how they reason about hypothetical scenarios. As a result, individuals may disagree not only on observed distributions but also on interventional and counterfactual ones.
Under this framework, we can describe how the two experts arrive at different classification probabilities by answering counterfactual questions, e.g.,~on a deferred instance. 
Our framework draws on foundational work in probabilistic and causal reasoning under uncertainty
\cite{Pearl1988ProbabilisticReasoning, PearlCausality2009}. We adopt a particular view of perception as disagreement in the interpretation of information conveyed through probabilities.

We define two kinds of causal perception: structural and parametrical. In structural perception individuals disagree on the cause-effect relationships, while in parametrical perception individual disagree on the causal effect between an established cause-effect pair.
We complement these two kinds of causal perception by providing an initial approach to model the processes by which individuals construct their SCM. Further, we discuss areas of interest within fair ML and point at promising directions of future work involving perception, such as situated bias and reformulating sensitive attributes as loaded attributes.
To showcase our framework, we present a series of examples involving ML applications and individual experts.

In the rest of the paper,
we introduce the necessary background in Section~\ref{sec:Introduction.Background}.
We then define probabilistic perception, or perception, in Section~\ref{sec:ProbabilisticPerception} and causal probabilistic perception, or causal perception, in Section~\ref{sec:CausalPerception}.
We present a first approach to the construction of the individual SCMs behind causal perception in Section~\ref{sec:TheFramework}.
We discuss the fair ML implications in Section~\ref{sec:Fairness}.
We conclude in Section~\ref{sec:Discussion}.

\subsection{Related Work}
\label{sec:Introduction.RW}

Perception has been studied mainly by psychologist, with views on it varying within the field.
We view perception as individual differences in interpreting the same information shaped by mental heuristics and expressed as probabilities; we discuss our choice later in Section~\ref{sec:ProbabilisticPerception}.
The focus within such view has been on the representativeness heuristic, in which an event is made to be more representative of a class than what it actually is, as measured by a higher probability \cite{tversky_judgment_1974, tversky_judgments_1981, tversky_extensional_1983}.
Bayesian modeling---in which the baseline representativeness of an event (the prior) is adjusted by the agent based on her experience (the posterior)---remains the common approach for modeling explicitly the representativeness heuristic and, thus, implicitly perception \cite{Costello2009HowPT, Bordalo2016_Stereotypes, Tentori22_CFonReasoning}.
Such view belongs to a cognitive-bias-driven interpretation of perception in which perception influences rational decision-making under uncertainty \cite{Kahneman2011Thinking}.
There are, for instance, constructivist \cite{Gregory1970IntelligentEye} and sensorial \cite{Goldstein1996SensationAndPerception} views that study the interpretative aspect of perception.
We are the first to use SCMs under this view of perception, which allows to frame it also in terms of hypothetical distributions that represent counterfactual reasoning. 
Further, we extend previous works since SCMs, or causal Bayesian networks \cite{PearlCausality2009}, include Bayesian reasoning.
The proposed causal perception framework, thus, can be viewed as a comparison of two individuals' posterior judgments and the causal theories that underlie those judgments.

ML researchers are increasingly interested in cognitive biases (see, e.g., \citet{BengioNIPS2019, Booch2021_ThinkingAI}).
Such works focus on how to create ML-driven intelligent systems that improve over and potentially replace the irrational human expert.
It is an active and interdisciplinary line of research (see, e.g., \citet{Agrawal2019EconomicsofAI}).
Our work adds to this growing line of research by making use of causal reasoning as a basis for modeling human-like reasoning \cite{Scholkopf2019_CausalML, PearlCausality2009}.
Further, there are narrower lines of ML work that study these cognitive biases and their impact on the human user of the ML application.
We highlight those studying the human-in-the-loop problem from a fairness and accuracy perspective (see, e.g., \citet{DBLP:conf/aaai/DeKGG20, DBLP:conf/aistats/MozannarLWSDS23, DBLP:journals/corr/abs-2405-18902}), which includes LtD \cite{DBLP:conf/nips/MadrasPZ18} and xAI \cite{DBLP:journals/pami/RongLNFQUSKK24}. 
Such works consider the human's interaction with the ML model, incorporating it within the problem formulation.
With some exceptions \cite{DBLP:journals/interactions/CarabanK20, DBLP:journals/pacmhci/RastogiZWVDT22, Yang2023Inner}, these works, however, do not give agency to the human user and formulate it as an additional and costly decision-maker, ignoring any influence from individual experience.
Different from these works, we explore the setting of having multiple users that disagree on the information provided. In that sense, our work gives more agency to the user by recognizing the role of perception among a set of heterogeneous users.

The role of these cognitive biases in fair ML is largely unexplored, with some exceptions.
% \cite{DBLP:conf/chi/EchterhoffYM22, DBLP:conf/aies/BertrandBEM22}.
Some of these works show how the cognitive biases can be exacerbated by the xAI techniques used by the expert to explain a model \cite{DBLP:conf/aies/BertrandBEM22} and can affect the expert's evaluation of a model's output \cite{DBLP:conf/chi/EchterhoffYM22}.
Further, given the problem of context-aware fairness, where we recognize that fairness has different meanings across humans, works like \citet{DBLP:conf/kdd/SrivastavaHK19} and \citet{DBLP:conf/aies/Yaghini0H21} design user experiments to test for the human perception of fairness.
These experiments show that the fairness of an outcome can be judged differently depending on who is the individual judging.
We add to these works by formalizing perception itself through causal reasoning.
In Section~\ref{sec:addRW} we further discuss perception within the causal fair ML literature.

\subsection{A Running Example}
\label{sec:Introduction.Example}

Throughout the paper we use a series of ML-enabled decision flow scenarios based on Example~\ref{ex:ControversialExampleUpdated}.

\begin{example}(College Admissions)
\label{ex:ControversialExampleUpdated}
    An admissions officer (the decision maker or DM) chooses the incoming class based on the applicants' profiles. 
    Assume a decision flow in which the officer admits, $Y=1$, or rejects, $Y=0$, applicants based on their SAT results, $X_1$, high-school GPA, $X_2$, and suitability scores $f(X_1, X_2)=G \in [0, 1]$ where $f$ is a ML model trained by the college.
    The officer, through the applicants' motivation letters, also has access to their address, $Z$.
    Consider three scenarios in which the officer relies on $f$ with varying degrees for the decision:\\
    \textit{(i)} $f$ abstains from classifying an applicant on which it is not confident and the officer must classify this applicant;\\
    \textit{(ii)} $f$ provides the same score for two applicants and the officer must choose one between these two applicants;\\
    \textit{(iii)} $f$ alone derives $Y$ using $G$ to rank applicants and admits the top-$k$ ones where $k$ is set by the officer.
\end{example}

The above example is an extended version of \citet{Kleinberg2019DiscAgeOfAlgo}'s college admissions tiebreaker example.
Scenarios \textit{(i)} and \textit{(ii)} represent a partially automated decision flow in which $f$ aids the DM, while scenario \textit{(iii)} represents a fully automated decision flow in which $f$ replaces the DM.
These are high-level but common ML-enabled scenarios.
Additional context is provided for each scenario as we extend Example~\ref{ex:ControversialExampleUpdated} moving forward.

%
% EOF
%

%% file: sections/2_Background.tex
In this work, we represent information as a set of $p$ discrete random variables $\mathbf{X} = X_1, \dots, X_p$. 
Let $P(\mathbf{X})$ denote the joint probability distribution of $\mathbf{X}$ with $P(\mathbf{x})$ representing the joint probability that $\mathbf{X}$ equals the $p$ instances $\mathbf{x} = x_1, \dots, x_p$. 
We treat individuals, like a human user, and objects, like a ML application, as agents. 

\paragraph{Senders and Receivers.}
Borrowing from signaling games \cite{Spence1973Signaling}, we consider two types of agents. 
A \textit{sender} $S \in \mathcal{S}$, with $\mathcal{S}$ denoting the set of senders, is an agent that provides information while a \textit{receiver} $R \in \mathcal{R}$, with $\mathcal{R}$ denoting the set of receivers, is an agent that interprets the provided information. 
We only borrow this distinction as our agents are non-strategic and engage in a single one-shot game, meaning that the sender sends and the receiver receives information once without tricking one another. For an overview of signaling games, see \citet{Sobel2020Signaling}.

\paragraph{Structural Causal Models.} 
A SCM~\cite{PearlCausality2009} describes the data-generating model of a process, allowing to reason how the $p$ random variables in $\mathbf{X}$ relate to each other as cause-effect pairs and how these pairs determine $P(\mathbf{X})$.
Formally, a SCM $\mathcal{M}$ is a tuple $\mathcal{M} = \langle \mathbf{U}, \mathbf{X}, \mathbf{F} \rangle$ that transforms a set of $p$ latent variables $\mathbf{U} \sim P(\mathbf{U})$ into a set of $p$ observed variables $\mathbf{X}$ according to a set of structural equations $\mathbf{F}$ such that:
\begin{equation}
\label{eq:SCM}
    P(\mathbf{U}) = P(U_1, \dots, U_j) \;\;\; X_j := f_j(X_{pa(j)}, U_j)
\end{equation}
for $j=1, \dots, p$ where $U_j \in \mathbf{U}$, $X_j \in \mathbf{X}$, and $f_j \in \mathbf{F}$. Each function $f_j$ maps the latent variable $U_j$ to the observed variable $X_j$ based on the subset of observed variables that directly cause $X_j$, or its causal parents $X_{pa(j)}$. 

The parental relations in a SCM induce a \textit{causal graph} $\mathcal{G}$, in which the nodes represent random variables and the directed edges between them causal relations.
We focus on acyclic graphs, meaning there are no loops in $\mathcal{G}$. This assumption turns $\mathcal{G}$ into a \textit{directed acyclical graph} (DAG) and ensures that information does not travel backwards, which is a common assumption \cite{Ruta2022_CausalDiscovery}.
% Further, we assume \textit{causal sufficiency}, another common assumption \cite{Ruta2022_CausalDiscovery}. It implies that $\mathbf{U}$ is independently distributed, meaning that are no are no unaccounted common causes. Both assumptions can be relaxed.

\paragraph{Linear SCMs.} 
% For readability, 
We focus on the \textit{additive noise models} (ANMs), where $f_j$ is a linear transformation:
\begin{equation}
\label{eq:AddSCM}
    f_j := \sum_{i=1}^{|pa(j)|} \beta_{ij} \cdot X_{pa(j)_i} + U_j
\end{equation}
with $\beta_{ij}\in\mathbb{R}$ denoting the \textit{causal weight} of the $i$-th parent $pa(j)_i$ of $X_j$. 
Our discussion applies to all SCMs, but we consider the ANM case for readability.

\paragraph{Causal Reasoning and Its Implied Distributions.}
The SCM $\mathcal{M}$ allows to reason about $P(\mathbf{X})$ in terms of observed and hypothetical scenarios.\footnote{\citet{Pearl2018WHy} present three levels of reasoning: observational (\textit{what is}), interventional (\textit{what if}), and counterfactual (\textit{what would have been if}). The latter two represent the hypothetical scenario. We use a simpler distinction.} For the observed scenario, or \textit{what is}, it is possible to disentangle the joint probability distribution $P(\mathbf{X})$ by factorizing it as a product of cause-effect pairs given the SCM $\mathcal{M}$:
\begin{equation}
\label{eq:MarkovCondition}
    P(\mathbf{X}) = \prod_{i=1}^{p} P\Big( X_i \, | \, X_{pa(i)} \Big)
\end{equation}
which simplifies reasoning about $P(\mathbf{X})$, as it states that $X_i$ is conditionally independent of its non-descendants given its parents $X_{pa(i)}$. This property is known as the \textit{Markovian condition} \cite{Peters2017}.

For the hypothetical scenarios, or \textit{what if}, it is possible to generate new distributions of $P(\mathbf{X})$ by intervening the SCM $\mathcal{M}$.
This is because SCMs build on an interventionist account of causality \cite{Woodward2005MakingThigsHappen}.
An intervention on a single variable $X_i$ is done via the \textit{do-operator}, $do(X_i=x_i)$, which replaces the structural equation in $\mathbf{F}$ for the variable $X_i$ with the value $x_i$. 
Interventions apply similarly for multiple variables, $do(X_i=x_i, X_j=x_j)$, replacing the structural equations for each variable individually.
Let $\mathcal{I}_{\mathbf{X}}$ denote the \textit{set of all interventions}, which is an index set with each index representing a specific intervention on the variables $\mathbf{X}$. We use $\emptyset \in \mathcal{I}_{\mathbf{X}}$ to denote the null intervention.
As \citet{DBLP:conf/uai/RubensteinWBMJG17} demonstrate,
$\mathcal{I}_\mathbf{X}$ has a \textit{natural partial ordering}, in which for interventions $i, j \in \mathcal{I}_\mathbf{X}, \, i \leq_\mathbf{X} j$ if and only if $i$ intervenes on a subset of the variables that $j$ intervenes on and sets them equal to the same values as $j$.
For instance, $do(X_i = x_i) \leq_\mathbf{X} do(X_i=x_i, X_j=x_j)$. It means the $j$ intervention can be done after the $i$ intervention without needing to change the modifications done by the $i$ intervention on the SCM $\mathcal{M}$.

Each intervention implies a well-defined joint distribution of  $\mathbf{X}$ variables $P(\mathbf{X})^{do(i)}$ for $i \in \mathcal{I}_{\mathbf{X}}$, called the $i$-interventional distribution.
Following \citet{DBLP:conf/uai/RubensteinWBMJG17}, 
we define the \textit{poset of all interventional distributions implied} by the SCM $\mathcal{M}$, where $\leq_{\mathbf{X}}$ is the natural partial ordering inherited from $\mathcal{I}_{\mathbf{X}}$, as:
\begin{equation}
\label{eq:PosetofDist}
    \mathcal{P}_{\mathbf{X}} := \Big( \Big\{ P(\mathbf{X})^{do(i)}: i \in \mathcal{I}_{\mathbf{X}} \Big\}, \, \leq_{\mathbf{X}} \Big).
\end{equation} 
By definition, $P(\mathbf{\mathbf{X}}) \in \mathcal{P}_{\mathbf{X}}$. 
Further, $\mathcal{P}_{\mathbf{X}}$ is a singleton comprised of $P(\mathbf{X})$ when $\mathcal{I}_{\mathbf{X}} = \{ \emptyset \}$. 
Furthermore, intuitively, $\mathcal{P}_{\mathbf{X}}$ represents all possible ways of reasoning causally about variables $\mathbf{X}$ as implied by a SCM $\mathcal{M}$.

%
% EOF
%

%% file: sections/3_ProbabilisticPerception.tex
We first present perception in its most basic form.
The goal is to formulate when two individuals, the receivers, interpret differently the same information from a ML application, formalized as the sender. 
We consider, in particular, the setting in which these individuals act as some sort of decision maker within a decision flow.

We deal with perception of probability as we assume that the individuals manifest their interpretations of the information received using probabilities.
This assumption is rooted in how humans reason under uncertainty, which has a long and established tradition across multiple fields that study bias in human decision-making (see, e.g., \citet{Kahneman2011Thinking, Thaler2008Nudge}).
We find this to be a reasonable assumption. 
Whether it is because the setting is inherently stochastic or the individual cannot possibly observe all the variables involved, there is always some degree of uncertainty involved in decision flows \cite{Pearl1988ProbabilisticReasoning}. 
Probabilistic reasoning helps capture the uncertainty behind making a choice based on a ML application's outcome.
Further, probability theory is one of the pillars of ML \cite{Goodfellow2016DeepLearning}. 
By describing perception in terms of probabilities, we can relate it to standard ML problem formulations.
Let us now offer a preliminary definition of perception of probability entirely based on this intuition.

\begin{definition}(Perception)
\label{def:Perception}
    For receivers $R_i, R_j \in \mathcal{R}$, given the information by sender $S \in \mathcal{S}$ in the form of a random variable
    % the instance $x$ of 
    % \textcolor{red}{the class} 
    $X$, perception occurs when, for a threshold $\epsilon \in \mathbb{R}^+$, we have that:
    \begin{equation}
    \label{eq:Perception}
        d \left( P_{R_i}(X), P_{R_j}(X) \right) > \epsilon 
    \end{equation}
    where the probability distributions $P_{R_i}(X)$ and $P_{R_j}(X)$ represent the interpretations by $R_i$ and $R_j$, respectively, of the information $X$, and $d(\cdot, \cdot)$ denotes a suitable distance metric between probability distributions.  
\end{definition}
\begin{remark}(From variable to variables)
    Definition~\ref{def:Perception} naturally applies to the $p$ random variables $\mathbf{X}$. In that case, $P_{R_i}(\mathbf{X})$ and $P_{R_j}(\mathbf{X})$ represent the joint probability distributions of the receivers for information $\mathbf{X}$.
    Further, we can consider the particular case in which one of the random variables denotes the target $Y$ and $P_{R_i}(Y|\mathbf{X})$ and $P_{R_j}(Y|\mathbf{X})$ represent the conditional joint probability distributions of the receivers for information $\mathbf{X}$ with $p-1$.
\end{remark}
\begin{remark}(Instance-based perception)
    Implicit to Definition~\ref{def:Perception} is a disagreement at the instance level $X=x$, meaning $d \big(P_{R_i}(x), P_{R_j}(x) \big ) > \epsilon$ where $P_{R_i}(x)$ and $P_{R_i}(x)$ are probabilities and $d(\cdot, \cdot)$ a suitable distance metric.\footnote{Here, the focus is on the specific instance $x$ as, indeed, distance among probability distributions in Equation~\ref{eq:Perception} may be $\leq \epsilon$, with still a number of instances having distance $> \epsilon$.}
    Intuitively, if a receiver can construct the probability distribution $P(X)$, then it can also compute the probability of any instance $x \in X$.
    It is also the case for the joint probabilities of the $p$ instances in $\mathbf{x}$ of the $p$ random variables in $\mathbf{X}$.
\end{remark}

With Definition~\ref{def:Perception}, we are interested in capturing how perception---as a disagreement in probability---would look like formally. 
We take for granted the process behind how these two receivers construct(ed) $P(X)$ and focus on the fact that there is a probabilistic disagreement between them.
% Further, 
The fact that there is disagreement of probability between $R_i$ and $R_j$ only has an impact when these two receivers must make a decision within the same decision-making context based on $P(X)$.
If $d \big( P_{R_i}(X), P_{R_j}(X) \big ) \leq \epsilon$, then we would not have a disagreement and we would be indifferent between $R_i$ and $R_j$. 
In that case, these two receivers, for our purposes, would be indistinguishable within said context.
Let us consider the next example. 

\begin{example}
\label{ex:ControversialExample_S1_ExtensionNum1}
    Let us consider Example~\ref{ex:ControversialExampleUpdated}, scenario \textit{(i)}.
    Suppose two admissions officers, $R_1$ and $R_2$, receive the same deferred applicant by the abstaining ML model $f$ with profile $\mathbf{x}= \langle x_1, x_2 \rangle$.
    Suppose that perception occurs as each officer constructs different $P(Y|X_1, X_2)$, such that $P_{R_1}(Y|x_1, x_2) > P_{R_2}(Y|x_1, x_2)$, for which officer $R_1$ accepts while $R_2$ rejects the applicant.
\end{example}

As Example~\ref{ex:ControversialExample_S1_ExtensionNum1} illustrates, Definition~\ref{def:Perception} dwells exclusively with the observational probability distribution or the \textit{what is}.
It does not allow for hypothetical scenarios, and the applicant is judged based on what is observed in his or her profile. 
There is no, for instance, counterfactual reasoning involved when deciding over the deferred applicant.
We expand such a setting in Section~\ref{sec:CausalPerception}.

\subsection{The Representativeness Heuristic}
% \subsection{The Probabilistic Problem of Representation}

Although Definition~\ref{def:Perception} cares only for the disagreement in probabilistic reasoning, we briefly discuss the mental heuristic commonly associated with perception as we view it in this work.
We do so to further support our focus on perception of probability.
The representativeness heuristic is one among many heuristics used by humans that lead to biased decision-making under uncertainty \cite{tversky_judgment_1974}.
Other heuristics, however, such as availability and anchoring, could also motivate our definition of perception.

% \citet{tversky_judgment_1974} argue that 
Probabilities can be understood as quantifying \textit{how representative} an instance $x$ is of a random variable $X$. 
Under this premise, for instance, Definition~\ref{def:Perception} reflects the setting in which two individuals can judge differently the representativeness of the same instance $x$ relative to the random variable $X$. 
Similarly, the \textit{representativeness heuristic} is used when humans assess how much one instance $x$ resembles another instance $x'$ that is known to belong to $X$, which may distort the estimation of $P(x)$. 
This is because, in such a case, reasoning centers on \textit{resemblance between instances}. The more one instance resembles the other instance, the more representative the former is of the latter instance. The question ``What is the probability that $x$ belongs to the random variable $X$?'' becomes ``To what degree the instance $x$ resembles the other known instance $x'$ in the random variable $X$?''. If the resemblance is high, then we judge the probability $P(x)$---read as $x$ belongs to $X$ or, equivalently, $x$ is generated by $X$---to be high.
Here, it helps overall to reason in terms of \textit{degree of representativeness}, which translates naturally into how we understand probabilities. 

\begin{definition}(Degree of Representativeness)
\label{def:DegreessRepr}
    For the random variable $X$ with known instance $x'$, the probability $P(x)$ is estimated by the degree of representativeness of $x$ as an instance of $X$ based on its resemblance to $x'$. 
    Given a distance $d(\cdot,\cdot)$ between instances, we have $P(x) \approx P(x')$ as $d(x, x') \approx 0$.
    This definition extends to the joint probability distribution of the collection of random variables $\mathbf{X}$, with instances $\mathbf{x}$ and known instances $\mathbf{x}'$.
\end{definition}

It follows from Definition~\ref{def:DegreessRepr} that, if the known instance $x'$ is viewed as \textit{representative of} $X$, denoted by a high-enough $P(x')$ within the relevant context, then $x$ is also representative of $X$ as captured by $P(x)$. 
The degree of representativeness is clearly a function of what someone considers representative of $X$ as captured by $x'$ and $P(x')$.
With respect to our framework for perception, it implies that receivers have their own $x'$ that they use to construct $P(x)$.

\begin{example}
\label{ex:ControversialExample_S1_ExtensionNum1_ctd}
    The occurrence of perception in Example~\ref{ex:ControversialExample_S1_ExtensionNum1} is driven by each admissions officer resorting to their own degree of representativeness when evaluating the deferred applicant with profile $\mathbf{x}= \langle x_1, x_2 \rangle$. Underlying the competing $P_{R_1}(Y|x_1, x_2)$ and $P_{R_2}(Y|x_1, x_2)$ is that each officer compares $\mathbf{x}$ relative to their own $\mathbf{x}'$.
\end{example}

For our purposes, we view this process as representing \textit{individual experience}.
It is private information known only to the receivers.
How such individual experience is constructed over time and elicited while reasoning by the receiver is beyond the scope of this work. 
We do, however, highlight another of Kahneman's work: \textit{norm theory} \cite{Kahneman1986_NormTheory}, which theorizes how individuals respond to the familiarity of an event by recruiting and creating alternative scenarios.
Under this theory, $R$ recruits a number of representations about an event. These representations are based on what $R$ views as a normal, with each scenario having a set of elements and each element having a set of features. These representations can be aggregated into a single scaled representation denoting the most common alternative, or \textit{the norm}, according to $R$. Such a process could help formulate under a ML approach how individual experience is constructed, stored, and used by the receiver.
We leave this for future work.

%
% EOF
%

%% file: sections/4_CausalPerception.tex
We now extend Definition~\ref{def:Perception} to account for probabilistic causal reasoning.
We want to capture disagreement between receivers beyond the \textit{what is} probability distribution by also considering the \textit{what if} probability distributions of the information sent by the sender.
We rely on a SCM~$\mathcal{M}$ to model the causal generative distribution~$P(\mathbf{X})$.
Notably, the single random variable case $P(X)$ is of no interest here as it lacks a parent or child to reason about.

\subsection{Why Causality?}

Our modeling choice is based on three factors.
First, under the premise that humans use probabilistic reasoning for decision-making under uncertainty \cite{Kahneman2011Thinking} as argued in Section~\ref{sec:ProbabilisticPerception}, the properties of a SCM $\mathcal{M}$ (Equation~\ref{eq:SCM}) offer a structured way to describe the information $\mathbf{X}$ as represented by $P(\mathbf{X})$. A SCM $\mathcal{M}$ allows to factorize $P(\mathbf{X})$ (Equation \ref{eq:MarkovCondition}) and to reason under hypothetical scenarios about $P(\mathbf{X})$ (Equation \ref{eq:PosetofDist}), in principle, similar to how humans interpret information and accumulate knowledge \cite{Woodward2005MakingThigsHappen, PearlCausality2009, Pearl2018WHy}.
Second, SCMs are increasingly used by ML researchers to approximate human-like reasoning \cite{Scholkopf2019_CausalML, Scholkopf_CausalRep_2021}. Given this trend, our proposed framework is compatible with the next wave of ML applications (see, e.g., \citet{DBLP:conf/nips/SteenkisteLSB19, DBLP:conf/iclr/DittadiTLWAWBS21}).  
Third, a SCM, in particular through its DAG $\mathcal{G}$, is a useful tools for engaging multiple stakeholders. A DAG, for instance, can be used to draw the assumptions about $\mathbf{X}$ and its data generating model in sensitive settings like discrimination testing \cite{CST23, DBLP:journals/jair/AlvarezR25} and synthetic data generation \cite{DBLP:conf/fat/BaumannCCIR23}. The DAG can essentially ``draw'' relevant individual experience.

We do not view SCMs as equivalent to human
reasoning. Rather, based on these three factors, we simply view SCMs as a useful and pragmatic tool for formalizing human reasoning
about $\mathbf{X}$ in terms of probabilities (similar, e.g., to \citet{DBLP:conf/icml/Loftus24}). 
We are aware that this is not a view widely held within the fairness community (see, e.g., \citet{Hu_facct_sex_20}). 
Future work could explore non-causal approaches to extend Definition~\ref{def:Perception} that also capture reasoning about hypothetical scenarios.

\subsection{Perception of Implied Probabilities}

\begin{definition}(Causal Perception)
\label{def:CausalPerception}
    For receivers $R_i, R_j \in \mathcal{R}$ with SCM $\mathcal{M}_{R_i}$ and $\mathcal{M}_{R_j}$ for the information $\mathbf{X}$ provided by sender $S \in \mathcal{S}$, causal perception occurs when for a threshold $\epsilon \in \mathbb{R}^+$:
    \begin{equation}
    \label{eq:CausalPerception}
        \overline{d} \left( \mathcal{P}_{\mathbf{X}_{R_i}}, \mathcal{P}_{\mathbf{X}_{R_j}} \right) > \epsilon
    \end{equation}
    where $\mathcal{P}_{\mathbf{X}_{R_i}}$ and $\mathcal{P}_{\mathbf{X}_{R_j}}$ represent the poset of all distributions of information $\mathbf{X}$ induced by $\mathcal{M}_{R_i}$ and $\mathcal{M}_{R_j}$, and $\overline{d}(\cdot,\cdot)$ denotes a suitable aggregated distance measure between two sets of probability distributions.
\end{definition}

The Definition~\ref{def:CausalPerception} extends Definition~\ref{def:Perception} into the realm of probabilistic causal reasoning.
It defines perception beyond a disagreement on the representation of $\mathbf{X}$ in terms of probabilities by also accounting for \emph{any} disagreement on the interventional distributions of the causal models.
In Example~\ref{ex:ControversialExampleUpdated}, for instance, we would go from disagreeing on ``What is the probability of an applicant being successful?'' to also disagreeing on ``What is the probability of an applicant being successful had the applicant been different on one of its variables?''.
The latter kind of disagreement is inherently linked to the SCM of each receiver.

For Equation~\eqref{eq:CausalPerception}, we propose in practice to compute the distance as an aggregation of these distributions, such as an average or a maximum (see, e.g., \citet{DBLP:journals/kais/GoldenbergW19}). 
Formally, given a set of interventions $\mathcal{I}_{\mathbf{X}}$, we consider the \textit{aggregated perception} for $R_i$ and $R_j$ in $\mathcal{R}$ as:
\begin{equation}
\label{eq:CausalPerceptionSpecific}
    \bar{d}(
    \mathcal{P}_{\mathbf{X}_{R_i}},
    \mathcal{P}_{\mathbf{X}_{R_j}}
    )
    =
    \Phi_{l \in \mathcal{I}_{\mathbf{X}}}
    d \left(
    P^{do(l)}_{R_i}(\mathbf{X}),
    P^{do(l)}_{R_j}(\mathbf{X})
    \right)
\end{equation}
such that $P^{do(l)}_{R_i}(\mathbf{X})$ denotes the \textit{l}-interventional distribution in the SCM $\mathcal{M}_{R_i}$ of receiver $R_i$ after performing the \textit{l}-intervention in $\mathcal{I}_{\mathbf{X}}$. 
The same applies for receiver $R_j$ and its SCM $\mathcal{M}_{R_j}$.
The choice of the distance function $d(\cdot, \cdot)$ and the aggregation function $\Phi$ in Equation~\eqref{eq:CausalPerceptionSpecific} will depend on the context and the kind of disagreement we wish to capture.
We could, for instance, define $d(\cdot,\cdot)$ as the Kullback–Leibler divergence and $\Phi$ as the average.

\begin{remark}(Specific interventional distributions)
    Notably, with Equation~\ref{eq:CausalPerceptionSpecific} we can also move away from ``any disagreement'' to ``a specific disagreement'' in terms of interventional distributions between receivers.
    In the special case in which no intervention occurs, or $\mathcal{I}_{\mathbf{X}}= \{ \emptyset \}$, Equation~\eqref{eq:CausalPerceptionSpecific} boils down to Equation~\eqref{eq:Perception}.
\end{remark}

Following up the previous remark, 
in the case of a specific \textit{l}-intervention, or $\mathcal{I}_{\mathbf{X}} = \{l\}$, 
Definition~\ref{def:CausalPerception} becomes:
\begin{equation}
\label{eq:Perception2}
    d \left( P_{R_i}(X)^{do(l)}, P_{R_j}(X)^{do(l)} \right) > \epsilon. 
    \end{equation}
Equation~\eqref{eq:Perception2} highlights how Definitions \ref{def:Perception} and \ref{def:CausalPerception} relate to each other as the difference in, respectively, the observational and the \textit{l}-interventional distributions.
Here, we study the disagreement of the interventional distribution resulting from the \textit{l}-intervention. 
We can generate these interventional distributions via the SCMs $\mathcal{M}_{R_i}$ and $\mathcal{M}_{R_j}$.

\subsection{Structural and Parametrical Perception}

Central to Definition~\ref{def:CausalPerception} are the SCMs of each receiver for $P(\mathbf{X})$.
If we assume that receivers draw all variables $\mathbf{X}$ using a SCM $\mathcal{M}$, then two kinds of causal perception emerge: structural and parametrical.
This assumption implies that the receivers, who receive the same information $\mathbf{X}$ from the sender, use all elements in $\mathbf{X}$, meaning each $X \in \mathbf{X}$ is also a variable in the SCM $\mathcal{M}$.
We consider the particular case of ANM (Equation~\ref{eq:AddSCM}) for illustrative purposes.

\begin{definition}(Structural Causal Perception)
\label{def:structural}
    It occurs when receivers $R_i, R_j \in \mathcal{R}$ disagree on the cause-effect pairs in, respectively, $\mathcal{M}_{R_i}$ and $\mathcal{M}_{R_j}$ for $P(\mathbf{X})$. It implies distinct DAGs $\mathcal{G}_{R_i}$ and $\mathcal{G}_{R_j}$.
\end{definition}
\begin{definition}(Parametrical Causal Perception)
\label{def:parametrical}
    It occurs when receivers $R_i, R_j \in \mathcal{R}$ agree on the cause-effect pairs for $P(\mathbf{X})$, meaning the their DAGs $\mathcal{G}_{R_i}$ and $\mathcal{G}_{R_j}$ are the same, but disagree on the %causal weights. It implies distinct 
    sets of structural equations $\mathbf{F}$ in respectively, $\mathcal{M}_{R_i}$ and $\mathcal{M}_{R_j}$. %as the same cause-effect pairs may have different parametric specifications.
\end{definition}

For instance, 
let $\mathbf{X} = \{X_1, X_2 \}$. 
In structural causal perception, $R_i$ draws $X_1 \rightarrow X_2$ while $R_j$ draws $X_2 \rightarrow X_1$. The distinct DAGs imply structurally different equations in the SCM as cause-effect pairs are reversed. 
In parametrical causal perception, assume $R_i$ and $R_j$ agree on $X_1 \rightarrow X_2$. Broadly, it means both will have a structural equation of the form $X_2 := \beta \times X_1 + U$ in their SCM. However, for $R_i$ $\beta > 0$ while for $R_j$ $\beta < 0$.
These two equations are parametrized differently coefficient-wise.

\begin{example}
\label{ex:ExampleForCausalPerceptions}
    Let us consider Example~\ref{ex:ControversialExampleUpdated}, scenario \textit{(ii)}.
    Suppose two admissions officers, $R_1$ and $R_2$, are tasked to choose independently which of the two applicants tied by the score $G$ from the ML model $f$ is admitted.
    Two additional points to consider. 
    First, the fact that $f$ gives the same score to both applicants implies comparable profiles, which can be summarized by a single $\mathbf{x}= \langle x_1, x_2 \rangle$. These two applicants are indistinguishable to $f$.
    Second, this fact allows the officers to derive the decision $Y$ without $G$. They can now consider other aspects of the applicants' profiles.
    Further suppose the officers resort to the applicants' zip code $Z$ for the tiebreaker. In particular, both officers reason about $P(Y|X_1, X_2, Z)$ by asking ``What would have been the score of an applicant had she or he been from the $Z=z$ zip code?'' 
    We can represent such hypothetical question using the $do(Z:=z)$ on the SCMs $\mathcal{M}_{R_1}$ and $\mathcal{M}_{R_2}$ of each officer, thus, looking at the interventional joint probability distribution that is $P(Y|X_1, X_2, do(Z:=z))$.
    
    Let us first consider the case of structural perception. 
    Suppose two competing DAGs for $\{Y, X_1, X_2, Z\}$:
    let $G_{R_1} = \{Z \rightarrow X_1, Z \rightarrow X_2, X_2 \rightarrow X_1, X_1 \rightarrow Y, X_2 \rightarrow Y \}$ and $G_{R_2} = \{Z \rightarrow X_2, X_2 \rightarrow X_1, X_1 \rightarrow Y, X_2 \rightarrow Y \}$, meaning $R_2$ does not agree with $R_1$ that $Z \rightarrow X_1$.
    Given these DAGs, each office factorizes the interventional joint probability distribution of interest differently when answering the what if question.
    We have $P(Y|X_1, X_2)P(X_2|Z)P(X_1|X_2, Z)P(Z)$ under $\mathcal{G}_{R_1}$ and $P(Y|X_1, X_2)P(X_2|Z)P(X_1|X_2)P(Z)$ under $\mathcal{G}_{R_2}$.
    The following set of structural equations corresponds to  $\mathcal{G}_{R_1}$:
    \begin{align}
        Z &:= U_1 \\
        X_1 &:= \beta_1 \cdot X_2 + \beta_2 \cdot Z + U_2 \\
        X_2 &:= \beta_3 \cdot Z + U_3 \\
        Y &:= \beta_4 \cdot X_1 + \beta_5 \cdot X_2 + U_4 
    \end{align}
    and a similar set of structural equations corresponds to $\mathcal{G}_{R_2}$ with the key difference that (10) becomes:
    \begin{align}
        % Z &:= U_1 \\
        X_1 &:= \beta_1 \cdot X_2 + U_2
        % X_2 &:= \beta_3 \cdot Z + U_3 \\
        % G &:= \beta_4 \cdot X_1 + \beta_5 \cdot X_2 + U_4 
    \end{align}
    as we exclude the term $\beta_2 \cdot Z$ since there is no $Z \rightarrow X_1$ cause-effect pair in $R_2$'s DAG.
    When reasoning about $Z$ and its impact on the other variables, each officer considers distinct interventional distributions due to the different factorizations of $P(\mathbf{X})$ as well as the competing set of structural equations. 
    $R_1$ constructs $P(Y|X_1, X_2)P(X_2|z)P(X_1|X_2, z)P(z)$ while $R_2$ $P(Y|X_1, X_2)P(X_2|z)P(X_1|X_2)P(z)$.
    Similarly, in terms of the set of structural equations, when substituting each $Z$ for the value $z$ using the \textit{do-operator} for $R_1$ we obtain:
    \begin{equation}
        Y:= \beta_4 \cdot (\beta_1 \cdot X_2 + \beta_2 \cdot z + U_2) + \beta_5 \cdot(\beta_3 \cdot z + U_3) + U_4
    \end{equation}
    while for $R_2$ we obtain:
    \begin{equation}
        Y:= \beta_4 \cdot (\beta_1 \cdot X_2 + U_2) + \beta_5 \cdot(\beta_3 \cdot z + U_3) + U_4
    \end{equation}
    where the term $\beta_4 \cdot \beta_2 \cdot z$ is missing for $R_2$ with respect to $R_1$ when making the decision. Such a difference can lead to a different applicant chosen by each officer based on $Z$.
    
    Let us now consider the case of parametrical perception.
    Suppose $R_1$ and $R_2$ agree on the previous DAG $\mathcal{G}_{R_1}$. 
    However, for equation (10) suppose $R_1$ considers $\alpha_1$ and $R_2$ considers $\alpha_2$ as the causal weight of $Z \rightarrow X_1$ (what is now $\beta_2$) such that $\alpha_1 \gg \alpha_2$.
    The same reasoning applies as with causal perception when constructing the interventional distribution $P(Y|X_1, X_2, do(Z:=z))$.
    The difference is that both receivers arrive at (14) after intervening $Z$ with but each has a distinct $\beta_4 \cdot \beta_2 \cdot z$ term.
    This term is $\beta_4 \cdot \alpha_1 \cdot z$ for $R_1$ and $\beta_4 \cdot \alpha_2 \cdot z$ for $R_2$, where $\beta_4 \cdot \alpha_1 \cdot z > \beta_4 \cdot \alpha_2 \cdot z$.
    Such a difference once again can determine which candidate each officer chooses based on $Z$.
    
    The term in question in both kinds of causal perception represents a potentially lower or higher SAT score because of the applicants zip code (read, neighborhood composition). We do this on purpose as the link between wealth (captured by $Z$) and aptitude (captured by $X_1$) is a devise one and of interest \cite{Kleinberg2019DiscAgeOfAlgo}.
\end{example}

The central idea with the two kinds of causal perception is that they account for why the two receivers in Definition~\ref{def:CausalPerception} disagree on the poset of implied distributions from a SCM $\mathcal{M}$.
The disagreement, by definition, comes from each receiver's SCM: with Definition~\ref{def:structural} the disagreement stems from different graphical structures while with Definition~\ref{def:parametrical} from different model parametrizations.
These competing SCMs represent the individual experience of each receiver.
% Recall that 
Here we take for granted the construction and elicitation of these individual experiences.
Notably, though, while in Definition~\ref{def:Perception} we made use of the representativeness heuristic to better understand individual experience in a probabilistic way, there is a lack for an equivalent framework to support Definition~\ref{def:CausalPerception}.
Therefore, in the next section we propose a causal modeling framework that motivates a receiver's individual experience as a SCM $\mathcal{M}$.

%
% EOF
%

%% file: sections/5_TheFramework.tex
How can we describe the process through which a receiver $R \in \mathcal{R}$ constructs its SCM $\mathcal{M}_R$?
We treat this section as a first approach given the complexity of the task. 
% It is a conceptual section.

\subsection{Key Functionalities}

Two processes are central to this framework: categorization and signification.
% Both have their origins in the social sciences: 
While \textit{categorization} entails sorting instances or classes into categories,
\textit{signification} entails representing the social meanings of the categories describing the instances or classes of interest \cite{Loury2019RacialIneq}.
For our purposes, given $X_{i}, X_{j} \in \mathbf{X}$ categorization implies listing the items that describe $X_{i}$ and $X_{j}$ while signification implies specifying how these items relate $X_{i}$ and $X_{j}$.
Notably, in signifying any two variables in $\mathbf{X}$ we want to specify causal-like statements between them.

To illustrate these processes, for instance, if $X_1$ represents SAT scores, then categorizing $X_1$ could amount to listing items such as ``standardized testing'', ``tutoring'', and ``performance under pressure''.
Similarly, signifying $X_1$ relative to zip code $Z$ could amount to stating that ``household income'' and ``school district'' are positively related to ``standardized testing''. The items ``household income'' and ``school district'' are obtained from signifying $Z$.

\begin{definition}(Categorization)
\label{def:Categorization}
    Let $\Theta^{R}(X=x)$ denote the categorization of $X=x$ by receiver $R$ such that:
    \begin{equation*}
        \Theta^{R}(X=x) =  \{ \theta_1^X, \dots, \theta_n^X \}
    \end{equation*}
    where $\theta_i^X$ is the \textit{i-th} item or category that receiver $R$ associates to $X$. It is possible for $\Theta^{R}(X=x) = \emptyset$.
    
    We define the \textit{categorization set} as:
    \begin{equation}
    \label{eq:Categorization}
        \vartheta^{R}(\mathbf{X}=\mathbf{x}) = \big \{ \Theta^{R}(X_i=x_i) \big \}_{i=1}^p
    \end{equation}
    where $p=|\mathbf{X}|$. Each $R \in \mathcal{R}$ comes with its own categorization set. It is implied that $X$ is a variable in the SCM $\mathcal{M}$ used by $R$. For readability, we write $\Theta^{R}(X)$ and $\vartheta^{R}$.
\end{definition}
\begin{definition}(Signification)
\label{def:Signification}
    Let $\Phi^{R}(X_i, X_j)$ denote the signification of $X_i, X_j$ by receiver $R$ such that:
    \begin{align*}
        \Phi^{R}(X_i, X_j) &=
        \hat{\Phi}^{R}(\Theta^{R}(X_i), \Theta^{R}(X_j))
    \end{align*}
    is one of $X_i \overset{\beta}{\rightarrow} X_j$ or $X_j \overset{\beta}{\rightarrow} X_i$,
    which reads as  ``$X_i$ causes $X_j$ with effect $\beta$'' or vice versa.
    It is possible for $\Phi^{R}(X_i, X_j)=\emptyset$.
    The idea is that receiver $R$ evaluates the items describing the \textit{i-th} and \textit{j-th} variables, derived from its categorization set, and from (the intersection of) these items derives (or not, if the result is $\emptyset$) a causal statement between them that depends on the kind of causal perception.
    
    Under structural causal perception (Definition~\ref{def:structural}) the $\Phi^{R}(X_i, X_j)$ is causal statement on the cause-effect pair and the causal effect. 
    Under parametrical causal perception (Definition~\ref{def:parametrical}) the $\Phi^{R}(X_i, X_j)$ is causal statement only on the causal effect.
    We define the \textit{signification set} as:
    \begin{equation}
    \label{eq:Signification}
        \varphi^{R} = \Big \{ \big \{ \Phi^{R}(X_i, X_j) \big \}_{j \neq i, j=1}^{p} \Big \}_{i=1}^p
    \end{equation}
     where $p=|\mathbf{X}|$. Each $R \in \mathcal{R}$ comes with its own signification set that is based on its own categorization set. It is implied that $X_i, X_j$ are variables in the SCM $\mathcal{M}$ used by $R$.
\end{definition}

Implementation-wise,
Definition~\ref{def:Categorization} relates to knowledge representation formalisms, such as ontologies \cite{Gruber1995TowardPF} or knowledge graphs \cite{10.1145/3447772} where $X$ is the class or concept and $\Theta^{R}(X)$ its attributes or properties.
Categorization would then amount to building a semantic model (for an ontology) or a network of facts (for a knowledge graph) that $R$ has for $\mathbf{X}$.
Similarly, Definition~\ref{def:Signification} relates to methods that argue over represented knowledge, such as logic argumentation \cite{DBLP:conf/foiks/BesnardCM14} or relational learning \cite{DBLP:conf/sigmod/SalimiPKGRS20}, among other techniques for argumenting over knowledge.
Signification would then amount to arguing how two variables relate causally to each other based on the affinity between the items describing them.
Under structural causal perception the answer would be structural, and under parametrical causal perception the answer would be parametric.

These two definitions are first approaches to capture how the receiver $R$ derives its SCM $\mathcal{M}_R$. 
Together, $\vartheta^{R}$ and $\varphi^{R}$ motivate the construction of $\mathcal{M}_R$.
The general idea here is that these two high-level processes are to Definition~\ref{def:CausalPerception} what the representativeness heuristic is to Definition~\ref{def:Perception}.

\subsection{Illustrative Examples}

Let us showcase how two receivers might construct their corresponding SCMs through the processes of categorization and signification, leading to causal perception.

\begin{example}(Categorization)
\label{ex:Categorization}
    In Example~\ref{ex:ControversialExampleUpdated}, scenario \textit{(ii)},
    let us assume that the admissions officer $R_1$ breaks the tie between the two applicants with the same suitability score $G$ from the ML model $f$ using the applicants' SAT scores $X_1$, high-school GPA $X_2$, and address $Z$. Recall from Example~\ref{ex:ExampleForCausalPerceptions} that it is implied that these applicants have the same $X_1$ and $X_2$.
    We define 
    $\vartheta^{R_1}$ of applicants' $X_1$, $X_2$, and $Z$ as:
    \begin{align*}
        \vartheta^{R_1} 
        = &\{ \Theta^{R_1}(X_1), \Theta^{R_1}(X_2), \Theta^{R_1}(Z) \} \\
        = &\{ \{ \text{tutoring, expensive, performative} \}, \\
        &\; \; \{ \text{discipline, school funding, potential} \}, \\
        &\; \; \{ \text{family income, school district} \} \}.
    \end{align*}
    The first line states the variables $R_1$ categorizes and the second line the descriptors for each variable according to $R_1$.
    Additionally, for later use in the upcoming examples 
    we define a second admissions officer $R_2$ that has a similar categorization set to $R_1$ but with $\Theta^{R_2}(X_1) = \emptyset$, which means that: $\vartheta^{R_2} = \{ \Theta^{R_1}(X_2), \Theta^{R_1}(Z) \}$.
\end{example}
\begin{example}(Signification for Structural Perception) 
\label{ex:SignificationUnfaithfulPerception}
    Based on $\vartheta^{R_1}$ and $\vartheta^{R_2}$ from Example~\ref{ex:Categorization}, we define the corresponding signification sets below.
    Assume due to the college's bylaws that an admissions officer determines $Y$ using $X_1$ and $X_2$ only. 
    Hence, $X_1 \rightarrow Y$ and $X_2 \rightarrow Y$ are given and $Z \rightarrow Y$ is not allowed; these cause-effect pairs are provided to and shared by $R_1$ and $R_2$. 
    For $R_1$:
    \begin{align*}
        \varphi^{R_1} 
        = &
        \{ \Phi^{R_1}(Z, X_1), 
        \Phi^{R_1}(Z, X_2),
        \Phi^{R_1}(Z, Y), \\
        & \; \; \Phi^{R_1}(X_1, X_2),
        \Phi^{R_1}(X_1, Y),
        \Phi^{R_1}(X_2, Y)
        \} \\
        = &
        \{ 
        \{\hat{\Phi}^{R_1}(\{\text{family income, school district} \}, \\ & \;\; \; \{\text{tutoring, expensive, performative} \}), \\
        & \; \; \hat{\Phi}^{R_1}(\{ \text{family income, school district} \},  \\ & \; \; \; \{ \text{discipline, school funding, potential} \}),
        \emptyset, \\
        & \; \; \hat{\Phi}^{R_1}(\{ \text{tutoring, expensive, performative} \},  \\ & \; \; \; \{ \text{discipline, school funding, potential} \} \}),\\
        & \; \; X_1 \rightarrow Y, X_2 \rightarrow Y \}\\ 
%        & \} \\
        = &\{ 
        Z \rightarrow X_1,
        Z \rightarrow X_2,
        X_2 \rightarrow X_1,
        X_1 \rightarrow Y,
        X_2 \rightarrow Y \}
    \end{align*}
    where the first equality states the pair of variables $R_1$ signifies; the second the items of each variable within each pair; and the third the cause-effect ordering for each pair based on these descriptors.
    For readability, we omit the $\beta$'s superscripts in the final line.
    For $R_2$ we have a similar signification set $\varphi^{R_2}$ with the exception of $\Phi^{R_2}(Z, X_1) = \emptyset$ as $\Theta^{R_2}(X_1) = \emptyset$.
    This results in the distinct DAGs discussed in  Example~\ref{ex:ExampleForCausalPerceptions} for structural causal perception.
\end{example}
\begin{example}(Signification for Parametrical Perception) 
\label{ex:SignificationFaithfulPerception}
    Similar to Example~\ref{ex:SignificationUnfaithfulPerception}, consider scenario \textit{(ii)} from Example~\ref{ex:ControversialExampleUpdated}.
    Suppose $R_1$ and $R_2$ now agree on $\mathcal{G}_{R_1}$.
    We formalize the scenario where two admissions officers consider the potential of the same applicant differently.
    Assume both officers have similar signification sets $\varphi^{R_1}$ and $\varphi^{R_2}$, except for $Z \rightarrow X_1$:
    \begin{align*}
         \Phi^{R_1}(Z, X_1)  
        & = \hat{\Phi}^{R_1}(\{ \text{family income, school district} \}, \\
        & \; \; \; \; \; \; \; \{ \text{tutoring, expensive, performative} \}) \\
        & = Z \overset{\beta_1}{\rightarrow} X_1\\
         \Phi^{R_2}(Z, X_1) 
        & = \hat{\Phi}^{R_1}(\{ \text{family income, school district} \}, \\
        & \; \; \; \; \; \; \; \{ \text{expensive, performative} \}) \\
        & = Z \overset{\beta_2}{\rightarrow} X_1
    \end{align*}
    such that $\beta_1 > \beta_2$. 
    % and where (for both receivers) the first line states the categorization of the $Z \rightarrow X_1$ cause-effect pair and the second 
    % the associations between descriptors with weights $\overline{\phi}(.)$ and the exact transformation $\tau_R(.)$ that aggregates them into 
    %the causal effect $\beta$ resulting from the intersection of sets. 
    % The $\beta_1$ and $\beta_2$ are conceived as illustrated in Figure~\ref{fig:FaithfulPerception}.
    This results in the distinct causal weights discussed in  Example~\ref{ex:ExampleForCausalPerceptions} for parametric causal perception.
\end{example}

In all three examples the address $Z$ acts as a proxy for socioeconomic background. 
This is intentional as our analysis is based on the original example from \citet{Kleinberg2019DiscAgeOfAlgo} in which an admissions officer decides between two applicants with the same SAT score but from different neighborhoods. 
We kept the ``controversy'' from the original example as it shows how a variable like $Z$ could condition an officer's decision-making based on how she reasons about $Z$ conditional on her individual experience.

\subsection{Additional Related Work}
\label{sec:addRW}

We extend Section~\ref{sec:Introduction.RW} by discussing causal perception within the fair causal ML literature.

\paragraph{Abstractions, Categorization, and Signification.}
Given a SCM $\mathcal{M}$ and its DAG $\mathcal{G}$ for $P(\mathbf{X})$, it is possible to imagine movements between low (or micro) and high (or macro) levels of causal abstraction.
Works on causal abstraction \cite{DBLP:conf/uai/RubensteinWBMJG17, DBLP:conf/uai/BeckersEH19, DBLP:conf/clear2/MassiddaGIB23} study how causal reasoning materializes and is preserved between levels. These works focus mainly on the notion of consistency, studying whether the how we reason on one level is consistent with the how we reason on another level given a SCM $\mathcal{M}$.\footnote{See \citet[Definition.~3; Theorem.~6]{DBLP:conf/uai/RubensteinWBMJG17} and \citet[Definition~3.1]{DBLP:conf/uai/BeckersEH19} for details.}
For example, suppose that we have one SCM describing the flow of water particles in a river and another SCM describing the flow of the river itself: consistency implies that intervening the water particles toward a specific direction should be equivalent to intervening the river toward that same direction.

With the processes of categorization (Definition~\ref{def:Categorization}) and signification (Definition~\ref{def:Signification}), we point at the concept of causal abstraction by allowing the receivers to go from the random variable $X$ (a high-level representation) to its \textit{n} descriptors $\{ \theta_1^X, \cdots, \theta_n^X \}$ (a low-level representation) and relate these to another variable's descriptors.
We do so by assuming a basic structure for $X$ and without formalizing the causal reasoning that takes place in $\mathcal{M}$ and $\mathcal{G}$ across these two levels of abstraction.
Our focus is conceptual and centered on knowledge representation, but we share the goal of formalizing how an agent's causal reasoning is impacted by the possible abstractions comprised in a SCM.

\citet{Hu_facct_sex_20} question whether a SCM can capture the meaning behind a sensitive random variable, such as gender, by formalizing it as a single node in a DAG.
Their work does not discuss causal abstraction, which should account for this critique, but notably argues for a molecular structure to sensitive attributes, such that there is gender (the high-level representation) and more granular nodes related to it (the low-level representation) in the DAG that simultaneously define gender as concept.
Both categorization and signification are inspired by \citet{Hu_facct_sex_20}'s discussion.
\citet{Mosse2025ModelingDisc} provide a similar conceptual discussion based solely on causal abstractions and focused on discrimination testing.
We are unaware of other fair causal ML works that discuss different levels of causal reasoning. 
We add to this line of work by considering how multiple agents can disagree on the abstraction levels when looking at the same (sensitive) attribute and how that leads to perception.

\paragraph{Counterfactual Fairness and Colliding Worlds.}
Given a SCM $\mathcal{M}$ and its DAG $\mathcal{G}$, counterfactual fairness (CF) \cite{Kusner2017CF} establishes that the observed outcome should be the same as the counterfactual outcome when intervening the sensitive attribute.
It remains the leading causal fairness metric \cite{Makhlouf2020_CausalFairnessSurvey}. By definition, these two outcomes belong, respectively, to the observational and counterfactual distributions that are both included in $\mathcal{P}_{\mathbf{X}}$~\eqref{eq:PosetofDist}. Extensions to CF, such as path-specific CF \cite{DBLP:conf/aaai/Chiappa19}, also deal with Equation~\eqref{eq:PosetofDist} as these always compare the observed distribution with respect to some interventional distribution given $\mathcal{M}$ and $\mathcal{G}$.

CF is with respect to a single SCM. 
% Hence, 
In a companion paper to \citet{Kusner2017CF}, \citet{DBLP:conf/nips/RussellKLS17} consider the robustness of CF when multiple world views, as in multiple SCMs describing $P(\mathbf{X})$, are used to compute CF.
Although multiple SCMs can be problematic as different SCMs lead to different CF results for the same $P(\mathbf{X})$ \cite{Ruta2022_CausalDiscovery},
\citet{DBLP:conf/nips/RussellKLS17} and later works (see, e.g., \citet{DBLP:conf/uai/KilbertusBKWS19} on hidden confounders) study the robustness of CF with a focus on the single SCM: all robustness claims are relative to a SCM of interest. 
The goal is to avoid disagreement by ``colliding'' the worldviews into one.

With causal perception we consider the opposite scenario from \citet{DBLP:conf/nips/RussellKLS17}.
We move away from colliding worlds by expecting disagreement among the receivers. 
Method-wise, we could indeed measure the robustness of a fairness metric like CF under causal perception (Definition~\ref{def:CausalPerception}) given these previous works. 
However, we diverge from them conceptually by considering the setting in which disagreement is allowed or even encouraged. 
To the best of our knowledge, what would amount to CF robustness in our setting remains unexplored.
Our causal perception definition is our contribution to this new line of work on alternative CF robustness claims. 
A similar argument applies to cases in which we rely on a SCM $\mathcal{M}$ and DAG $\mathcal{G}$ but may allow multiple perspectives from stakeholders, such as discrimination testing \cite{CST23, DBLP:journals/jair/AlvarezR25} and synthetic biased data generation \cite{DBLP:conf/fat/BaumannCCIR23}.

%
% EOF
%

%% file: sections/6_PerceptionAndFairness.tex
Finally, we explore three fair ML areas that benefit from the proposed formalization of (causal) perception.

\subsection{Situated Bias}
\label{sec:Fairness.Bias}

In previous sections, we avoided the term ``bias'' despite the connection between perception as a cognitive phenomenon and biased human decision-making \cite{Kahneman2011Thinking}.
This choice was motivated by two  considerations.
First, the notion of bias often associated to our chosen view of perception (recall, Section~\ref{sec:Introduction.RW}) does not align directly with the definition commonly used in algorithmic fairness. While the former refers to a departure from rational decision-making as described by basic probability laws \cite{Kahneman2011Thinking}, the latter refers to ``demographic disparities in algorithmic systems that are objectionable for
societal reason'' \cite{barocas-hardt-narayanan}.
The latter interpretation is usually the primary concern in fairness-related discourse.
Second, building on this distinction, the notion of bias in algorithmic fairness often carries a negative connotation as one aims to capture objectionable disparities. 
There is nothing, in principle, negative being captured by Definitions~\ref{def:Perception} and \ref{def:CausalPerception}.
Both definitions simply formalize a disagreement in probabilistic and causal probabilistic reasoning between two individuals.
We must, thus, be more precise by ``situating'' the bias \cite{Haraway1988SituatedKnowledges}.

The disagreement between receivers $R_i$ and $R_j$ may lead to bias, but to speak of the kind of bias that drives algorithmic unfairness we must not only have a disagreement but also a preference on which interpretation of the information $\mathbf{X}$ is preferred. 
Similar to how \citet{Haraway1988SituatedKnowledges} argues that all objective knowledge is based on a partial view of the problem,\footnote{Objectivism is a function of what we choose to see (situation), how we choose to see it (location), and from where we choose to see it (position), which are telling of privilege and power relations.} our framework can be used to situate bias by defining one receiver's (causal) probabilistic interpretation of $\mathbf{X}$ as the reference interpretation.
The underlying idea is that since individual experience, which we construct through a SCM $\mathcal{M}$, shapes decision-making, then we can use our formalization to explore settings in which we cannot reach an agreement on what is fair or want to be explicit on what fairness looks like within a specific context. 

\begin{example}
    In Examples ~\ref{ex:SignificationUnfaithfulPerception} and \ref{ex:SignificationFaithfulPerception} we ca define $R_1$ as the representative fair receiver given its reasoning behind $X_1$ and $Z$. $R_1$ would then represent the desired decision-maker for breaking potential ties between applicants in terms of the suitability score $G$ of the ML model $f$.
\end{example}

\subsection{Loaded Attributes}
\label{sec:Fairness.LoadedAttribute}

Similarly, in previous sections we did not discuss sensitive attributes, like gender or race, and their role in perception. 
We argue that these attributes are prone to induce perception, as they are summaries of complex historical and social processes \cite{Bonilla1997_RethinkingRace, Sen2016_RaceABundle}. As such, they are likely to affect how individuals are perceived by others.
We are referring, for example, to the conceptual difference between describing an individual as female versus feminine \cite{Hu_facct_sex_20}, both of which are based on the attribute gender. Female refers to a category of gender while feminine refers to a set of behavioral expectations attributed to females: i.e.,~the biological or phenotypic label versus the socially constructed identity.
We describe these attributes as loaded because they are likely to lead to different interpretations among receivers.
If $X \in \mathbf{X}$ is a loaded attribute, then it should be easier for a receiver $R$ to evoke individual experience about $X$.

% Broadly, 
We regard loaded attributes as attributes that thrive on stereotypes of social categories shared and maintained by the receivers.
A social category is the result of classifying people into groups over shared perceived identities~\cite{bowker1999sorting}. 
We refer to a social category as a social construct when the classification is also used purposely to enforce exclusionary policies~\cite{Mallon2007SocialConstruction}. 
Sensitive attributes are examples of social constructs.
A stereotype refers to the cognitive representation people develop about a particular social category, based around beliefs and expectations about probable behaviors, features and traits, which translate into implicit or explicit attitudes that materialize into bias \cite{Camiel_Stereotypes_2019, Johnson2020structurebias}.

The main difficulty, as discussed in Section~\ref{sec:addRW}, is how to model  causally such distinction and, further, how it translates into a potential disagreement among the decision makers.
The causal perception framework offers a way to formalize these overloaded attributes as elements of individual experience from the perspective of the receiver. 
The SCMs condition how receivers interpret information, which is particularly important given the potential biases they may hold toward certain social groups.

\begin{example}
\label{ex:ControversialExampleScen2} 
    As implied in Examples~\ref{ex:SignificationUnfaithfulPerception} and \ref{ex:SignificationFaithfulPerception},
    the admissions officers' consideration of the zip code, $Z$, of each applicant serves as an attempt to infer their socioeconomic background.
    By associating $Z$ with $X_1$ and $X_2$ and drawing on stereotypes (like correlations between wealth and SAT scores), an officer may develop biased inferences, which ultimately lead to favor one applicant over the other.
\end{example}

\subsection{The Human-in-the-Loop}
\label{sec:Fairness.Applications}

The next example, while centered on automated decision-making (ADM), illustrates the same interpretative challenges discussed in the previous ones for scenarios \textit{(i)} and \textit{(ii)}.
We view the admissions officer as one receiver and her supervisor as another receiver. 
A natural question to ask then is ``What happens if they disagree on the $k$ applicants selected by $f$?'' in scenario \textit{(iii)}.
Such situation highlights a broader issue with ML pipelines: the final decision often culminates in human oversight. 
From a fairness perspective, it is essential to develop frameworks that accommodate subjective interpretations and context-aware notions of fairness \cite{DBLP:conf/kdd/SrivastavaHK19, DBLP:conf/aies/Yaghini0H21, DBLP:conf/forc/0001KN0SW21}.

\begin{example}
\label{ex:ControversialExampleExtended}
    Consider scenario \textit{(iii)} in Example~\ref{ex:ControversialExampleUpdated}. Given the top-$k$ applicants by the ML model $f$,
    suppose that the admissions officer is asked by the college to explain for the $k$ chosen applicants by $f$. 
    The officer uses xAI tools, like feature importance methods, to understand the model's outcome and explain it to her supervisor.
    Assume that the supervisor also has access to $f$ and the same xAI tools.
    Suppose the officer and her supervisor interpret differently the feature importance explanations.
\end{example}

Example~\ref{ex:ControversialExampleExtended} is based on studies such as \citet{DBLP:conf/aies/BertrandBEM22} that consider the limitations of xAI methods under the threat of human cognitive biases.
% As studied by \citet{DBLP:conf/aies/BertrandBEM22}, suppose that in Example~\ref{ex:ControversialExampleExtended} the officer and her supervisor interpret differently the feature importance explanations. 
This divergence between officer and supervisor highlights a key challenge in xAI: how can we design explanations that minimize interpretive ambiguity among users? 
Similar to nudging in behavioral economics \cite{RBE-0092}, which was inspired by Tversky and Kahneman's work on cognitive biases \cite{Thaler2008Nudge}, future xAI tools should aim to anticipate a range of plausible user interpretations and proactively guide users toward a shared, intended understanding.
The causal perception framework allows to formalize such set of user-based interpretations in a way that is compatible with ML applications and xAI tools.

%
% EOF
%

%% file: sections/7_Discussion.tex
In this work, 
we treated perception as a disagreement between individual probabilistic interpretations of the same bit of information, and defined perception of probabilities and perception of implied probabilities.
For the latter, which we termed causal perception, we used structural causal models (SCMs) to formalize disagreement between observed and interventional (read, hypothetical) distributions.
Further, we formalized two kinds of causal perception based on when individuals disagree on the causal graph (structural perception) and when they disagree on the causal effect(s) given the same casual graph (parametrical perception). 
Furthermore, we proposed a first approach to how individuals construct these disagreeing individual SCMs through the processes of categorization and signification.

We also outlined fairness research areas that benefit from the proposed framework. 
Notably, causal perception is useful in contexts in which multiple interpretations and representations of information occur by experts.
It is, in turn, useful for problems involving diverse stakeholders that interact with the ML system.
The framework allows to formally position fairness in such a setting as well as to reformulate sensitive attributes as loaded with perceptions.

This paper is, above all, conceptual.
Next steps include, for instance, using ontologies \cite{Gruber1995TowardPF} and relational learning \cite{DBLP:conf/sigmod/SalimiPKGRS20}---with \citet{Kahneman1986_NormTheory}'s norm theory in mind---to construct in practice the SCMs that denote individual experience.
Revisiting causal perception under Pearl's three levels of reasoning (here, we simplified it to two levels) is also a next step \cite{DBLP:books/acm/22/BareinboimCII22}.
In general, future work should validate the framework empirically through case studies.
Additionally, causal perception assumes an explicit level of causal reasoning among individuals that may not be realistic as well as assumes SCMs as individual prior knowledge. 
Future work should evaluate how practical and scalable these individual SCMs are for the context in question. 

%
% EOF
%